\documentclass[twoside,twocolumn,10pt]{article}
\usepackage[bottom,marginal,stable,multiple]{footmisc}
\usepackage{wscg}
\usepackage{mathtools}
\usepackage{amsmath}
\usepackage{url}
\usepackage[pdftex]{graphicx}
\graphicspath{{figures/}{plots/}}
\usepackage{subcaption}
\usepackage{adjustbox}
\usepackage{nopageno}
\usepackage[pdftex]{color}
\usepackage[numbers]{natbib}
\usepackage{natbibspacing}
\usepackage{tabu}
\usepackage{booktabs}
\usepackage{import}
\usepackage{flushend}
\usepackage{hyperref}

\makeatletter
\def\affiliation#1{\def\@affiliation{#1}}
\def\@maketitle{%
  \begin{center}%
  \let \footnote \thanks
    \sffamily
    {\fontsize{16pt}{19.2pt} \bfseries \@title \par}%
    \vskip 1.0em%
    {%
      \lineskip .5em%
      \begin{tabular}[t]{c}%
        \@author
      \end{tabular}%
      \vskip 0.5em%
      \@affiliation%
      \par}%
  \end{center}%
  \par
  \vskip 0.5em}
\makeatother

\title{Compression Artifacts Removal Using\\ Convolutional Neural Networks}
\author{Pavel Svoboda \and Michal Hradis \and David Barina \and Pavel Zemcik}
\affiliation{Faculty of Information Technology\\Brno University of Technology\\Bozetechova 1/2, Brno\\Czech Republic\\\{isvoboda,ihradis,ibarina,zemcik\}@fit.vutbr.cz}

\def\TrimL{.1}
\def\TrimR{.3}

\begin{document}

\twocolumn[{\csname @twocolumnfalse\endcsname

\maketitle

\begin{center}
	\hspace*{\fill}
    \adjustbox{trim={\TrimL\width} {.4\height} {\TrimR\width} {.2\height},clip}
    {\includegraphics[width=.8\linewidth]{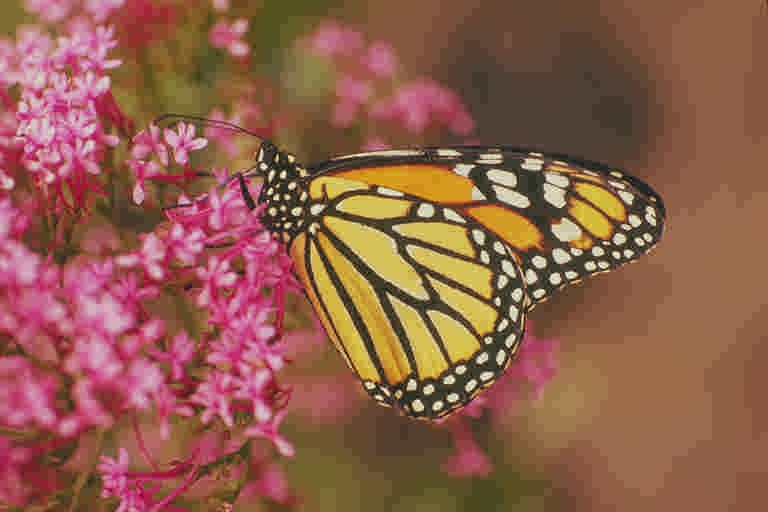}}
    \hspace*{\fill}
    \adjustbox{trim={\TrimL\width} {.4\height} {\TrimR\width} {.2\height},clip}
    {\includegraphics[width=.8\linewidth]{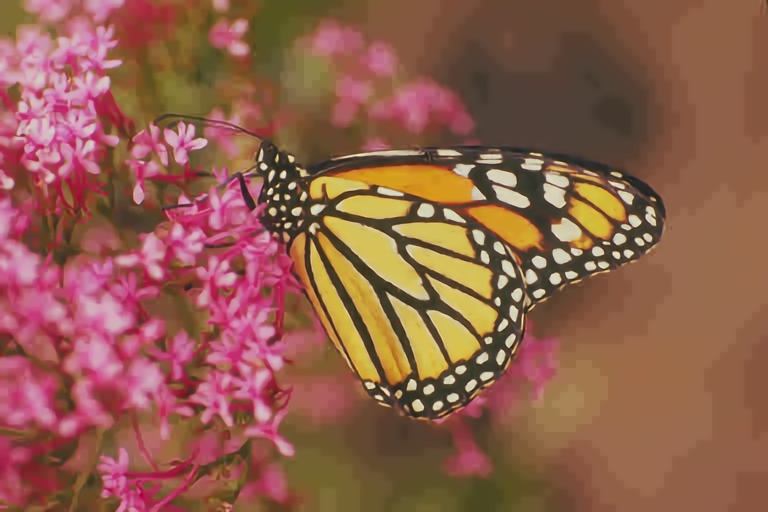}}
    \hspace*{\fill}
\end{center}

\begin{abstract}
This paper shows that it is possible to train large and deep convolutional neural networks (CNN) for JPEG compression artifacts reduction, and that such networks can provide significantly better reconstruction quality compared to previously used smaller networks as well as to any other state-of-the-art methods.
We were able to train networks with 8 layers in a single step and in relatively short time by combining residual learning, skip architecture, and symmetric weight initialization.
We provide further insights into convolution networks for JPEG artifact reduction by evaluating three different objectives, generalization with respect to training dataset size, and generalization with respect to JPEG quality level.
\end{abstract}

\subsection*{Keywords}
Deep learning, Convolutional neural networks, JPEG, Compression artifacts, Deblocking, Deringing

\vspace*{1.0\baselineskip}
}]

\section{Introduction}
\label{sec:Introduction}

\copyrightspace

This work presents a novel method of image restoration using convolutional networks that represents a significant advancement compared to the state-of-the-art methods.
We study the direct approach~\cite{Hradis2015} in which a fully convolutional network accepts a degraded image as input and outputs a high quality image.
By making a number of important improvements regarding the network architecture, initialization, and training, we are able to train large and deep networks for JPEG compression artifact reduction which surpass the state-of-the-art in this task.
The networks predict a residual image~\cite{Kim2015} describing changes to be applied to the input image, and they incorporate skip connections~\cite{Long2015} which allow information to bypass the middle layers.
We reduce the "saturation" of ReLU units in deeper layers by centering filters during network initialization which allows us to use significantly faster learning rates.

Lossy image compression achieves high compression ratios through elimination of information that does not contribute to human perception of images, or contributes as little as possible.
Due to the limitations of the human visual system, such loss of information may be acceptable in many scenarios but the introduced visual artifacts become unacceptable at higher compression ratios.
The primary methods currently used for lossy image compression include JPEG and JPEG 2000.
This paper focuses on the JPEG compression method~\cite{T.81} and the degradation it causes.
The JPEG compression chain consists of a block-based discrete cosine transform (DCT), followed by a quantization step utilizing a quantization matrix, and an entropy coding.
The decompression follows this process in reverse order.

Blocking, blurring, and ringing artifacts are typical examples of image degradation caused by the lossy compression methods.
Considering the JPEG method, the degradation is the result of information loss in the DCT coefficient quantization step.
More specifically, the blocking artifacts are caused by the grid segmentation into $8\times8$ cells employed in the JPEG standard and the resulting discontinuities at the cell edges.
The ringing artifacts (or the Gibbs phenomenon) are induced oscillations caused by removal of high frequencies during the quantization.
The removal of high frequencies causes blurring as well, but the blurring is less noticeable compared to the ringing artifacts.
Blocking is mostly noticeable in low-frequency regions, while the ringing artifacts are especially well noticeable around sharp edges.

The convolutional networks have to learn to recognize the compression artifacts and fill them appropriately with respect to the neighboring image content.
In this sense, the networks incorporate both the data term and prior regularization term of standard image restoration techniques, and they can make use of correlations between image content and the image degradation.

Convolutional networks have been successfully used in many image restoration tasks including super resolution~\cite{Dong2014,Kim2015}, denoising~\cite{Jain2009}, structured noise removal~\cite{Eigen2013}, non-blind deconvolution~\cite{Schuler2013, Xu2014}, blind deconvolution in specific image domains~\cite{Hradis2015,Svoboda2016}, and sub-tasks of blind deconvolution~\cite{Schuler2015}.
Our work was mostly inspired by the large deblurring networks of Hradis \textit{et al.}~\cite{Hradis2015}, and by Kim \textit{et al.}~\cite{Kim2015} who showed that residual learning together with good weight initialization enabled training of large convolutional networks for super resolution.
We extend the work of Dong \textit{et al.}~\cite{Dong2014} who achieved state-of-the-art compression artifact reduction even with very small convolutional networks.
However, they were not able to scale up their networks due to problems with training convergence.

\section{Related Work}
\label{sec:Related-Work}

A large number of methods designed to reduce compression artifacts exist ranging from relatively simple and fast hand-designed filters to fully probabilistic image restoration methods with complex priors~\cite{Wong2009} and methods which rely on advanced machine learning approaches~\cite{Dong2014}.

Simple deblocking and artifact removal postprocessing filters are included in most image and video viewing software.
For example, the FFmpeg framework includes the simple postprocessing (spp) filter~\cite{Nosratinia1999} which simply re-applies JPEG compression to the shifted versions of the already-compressed image, and averages the results.
The spp filter uses the quantization matrix (compression quality) of the original compressed image as the matrix has to be stored with the image to allow for decompression.
Pointwise Shape-Adaptive DCT (SA-DCT)~\cite{Foi2006,Foi2007}, in which the thresholded or attenuated transform coefficients are used to reconstruct a local estimate of the signal within the adaptive-shape support, is currently considered the state-of-the-art deblocking method.
However, similarly to other deblocking methods, SA-DCT overly smooths images and it is not bale to sharpen edges.
In video compression domain, advanced in-loop filters (deblocking and SAO filters) known from video compression standards like H.264 or H.265 are obligatorily applied.
A completely different deblocking approach was presented in \cite{Yang2000}, where the authors applied DCT-based lapped transform on the signal already in the DCT domain in order to undo the harm done by the DCT domain processing.
However, the video in-loop deblocking methods, SA-DCT deblocking (only to estimate parameters), and methods derived from the lapped DCT rely on the cognizance of the DCT grid.
Unlike these methods, the method proposed in this paper is able to process images without such knowledge.

This work focuses on application of convolutional networks to reconstruction of images corrupted by JPEG compression artifacts.
Convolutional networks belong to an extensively studied domain of deep learning~\cite{Bengio2009}.
Recent results in several machine learning tasks show that deep architectures are able to learn the high level abstractions necessary for a wide range of vision tasks including face recognition~\cite{Taigman2104}, object detection~\cite{Girshick2014}, scene classification~\cite{Krizhevsky2012}, pose estimation~\cite{Toshev2014}, image captioning~\cite{Vinyals2015}, and various image restoration tasks~\cite{Dong2014,Kim2015,Jain2009,Eigen2013,Schuler2013,Xu2014,Hradis2015,Svoboda2016,Schuler2015}.
Today, convolutional networks based approaches show the state-of-the-art results in many computer vision fields.

Small networks were historically used in image denoising and other tasks.
On the other hand, deep and large fully convolutional networks have become only recently important in this field.
Burger \textit{et al.}~\cite{Burger2012} used feed forward three layer neural network for image denoising.
While there were attempts to use neural networks for denoising before, Burger \textit{et al.} showed that this approach can produce state-of-the-art results when trained on a sufficiently large dataset.

A non-blind deconvolution method of Schuler \textit{et al.}~\cite{Schuler2013} uses a regularized inversion of the blur kernel in Fourier domain followed by a multi-layer perceptron (MLP) based denoising step.
The shortcoming of the approach is that a separate MLP models have to be trained for different blur kernels, as a general models trained for multiple blur kernels provide inferior reconstruction quality.
Schuler \textit{et al.}~\cite{Schuler2015} introduced a learning based approach to blind deconvolution.
They perform a regression from the blurred image towards the source blur kernel.
The neural network itself is trained to extract image features useful for estimation of the blur point spread function.
Sun \textit{et al.}~\cite{Sun2015} presented CNN-based approach for non-uniform motion blur removal which classified image patches into closed set of blur kernel types.
The local classification outputs were used as input to a Markov random field model which estimates the dense non-uniform motion blur field over the whole image.
Hradis \textit{et al.}~\cite{Hradis2015} trained CNNs composed of only convolutional layers and rectified linear units (ReLU) to directly map blurred and noisy images of text images to high quality clean images.
The approach was extended by Svoboda \textit{et al.}~\cite{Svoboda2016} who demonstrated high quality deblurring reconstructions for car license plates in a real-life traffic surveillance system.
Their results show that a single CNN can be trained for a full range of motion blurs expected to appear in a specific traffic surveillance camera resulting in a robust and fast system.

Dong \textit{et al.}~\cite{Dong2014} introduced super-resolution convolutional neural network (SRCNN) to deal with the ill-posed problem of super-resolution.
The SRCNN is designed according the classical sparse coding methods -- the three layers of SRCNN consist of feature extraction layer, a high dimensional mapping layer, and a final reconstruction layer.
The very deep CNN based super-resolution method proposed by Kim \textit{et al.}~\cite{Kim2015} builds on the work of Dong \textit{et al.}~\cite{Dong2014} and it shows that deep networks for super-resolution can be trained when proper guidelines are followed.
They initialized networks properly and they used so-called residual learning in which the network predicts how the input image should be changed instead of predicting the desired image directly.
Residual learning appears to be very important in super-resolution.
The resulting 20 layers deep networks trained with adjustable gradient clipping significantly outperform previous approaches.
However, it is unclear how effective residual learning would be in other image processing tasks where the networks inputs and outputs are not correlated that strongly as in super-resolution.
We follow this approach in our work on JPEG reconstruction.

Convolutional networks have previously been used for suppressing compression artifacts by Dong \textit{et al.}~\cite{Dong2015}, who proposed a compact and efficient CNN based on SRCNN -- artifacts removing convolutional network (AR-CNN).
AR-CNN extends the original architecture of SRCNN with feature enhancement layers.
The network training consist of two stages -- a shallow network is trained first and it is used as an initialization for a final 4 layer CNN.
As reported in the paper, this two stage approach improved results due to training difficulties encountered when training the full 4 layer network from scratch.
The authors also state that they aim to achieve feature enhancement instead of just making the CNN deeper.
They argue that although the deeper SRCNN introduces a better regressor between the low-level features and the reconstruction, the bottleneck lies on the features.
Thus the extracting layer is augmented by the enhancement layer which together may provide better feature extractor.

We adapt the idea of residual learning~\cite{Kim2015} for the JPEG compression artifact removal based on CNN.
We follow the assumption "deeper is better" and we try to learn our deep residual CNNs in a single step by creating a new recipe including initialization, network architecture, and high learning rates.
The resulting networks significantly outperform the classical JPEG compression artifact removal methods, as well as, the \mbox{AR-CNN}~\cite{Dong2015} on common dataset measured by PSNR, specialized deblocking assessment measure \mbox{PSNR-B}, and SSIM.

\section{CNN image enhancement}
\label{sec:CNN-image-enhancement}

In computer vision, CNNs are most extensively studied in the context of classification, semantic class segmentation, object detection, and captioning where the networks are often constrained to a fixed input size.
This is due to the fully connected layers which are used as the final layers in order to aggregate information from a whole image.
In low level image processing (but not limited to it), the so-called fully convolutional neural networks~\cite{Long2015} (FCN) are preferred as they behave as non-linear convolutional operators -- they process each image position the same way and they can be applied to images of arbitrary size.%
\footnote{In practice, the minimum size of processed images is constrained by the receptive field size of the network.}
The architecture of fully convolutional networks is limited to convolutional operations (linear convolution, so-called deconvolution, local response normalization, and local pooling) and element-wise operations.
Most image processing networks use only convolutions and element-wise non-linearities (ReLU, sigmoid, tanh)~\cite{Hradis2015,Svoboda2016,Dong2015,Kim2015,Dong2014,Schuler2013,Schuler2015}.
In the case that no pooling and no deconvolution layer is used, the size of the input is reduced only by size of the convolution layer kernels (by the size of receptive field).

The fully convolutional networks $F$ used in our work consist of an input data layer $F_0$, convolutional layers $F_{\ell}$, where
$0 < \ell \le L$ with $F_{\ell}$ weights represented as convolutional kernels $W_{\ell}$ with their \mbox{biases} $b_{\ell}$, and element-wise $\max$ operations (ReLU)
as follows:
\begin{align}
	\begin{split}
		F_{0}(y) &= y \\
		F_{\ell}(y) &= \max(0, W_{\ell} \ast F_{\ell-1}(y) + b_{\ell} )\\
		F(y) &= W_{L} * F_{L-1}(y)+b_{L}
	\end{split}
	\label{eq:FCN}
\end{align}
Where $y$ is the distorted input image and $F(y)$ is the restored output image.

We use the standard mean squared error (MSE) objective function 
\begin{align}
	\frac{1}{n} \sum_{i=1}^{n} \left\lVert F(y_{i}) - x_{i} \right\rVert_{2}^{2},
	\label{eq:MSE}
\end{align}
which is often used for general image enhancement. It is computed on a training data represented as pairs ${(y_{i}, x_{i})},$
$0 < i \le n$, where $y_{i}$ represents the reconstructed image and $x_{i}$ its corresponding clean image.

\begin{figure}
	\centering
	\includegraphics[width=\linewidth]{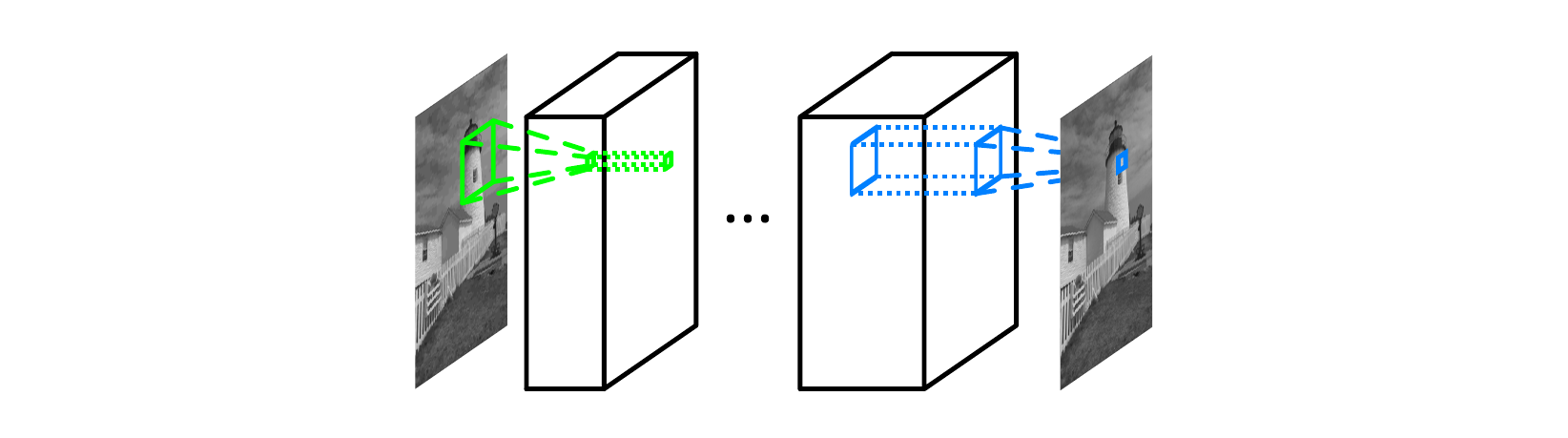}
	\caption{Illustration of a network with direct architecture.}
	\label{fig:schema-direct}
\end{figure}

\medskip
\subparagraph{Direct mapping objective.}
In direct mapping shown in Figure \ref{fig:schema-direct}, the networks learn to transform corrupted images directly to clean images.
This approach leads to high quality results in specific low level image processing tasks i.e. in blind and non-blind deconvolution for text denoising or motion deblurring~\cite{Hradis2015, Svoboda2016}, in super-resolution \cite{Dong2014} or JPEG compression artifacts reduction~\cite{Dong2015}.
Direct mapping forces the network to transfer the whole image through all its layers until it reaches the output.
The learning of such autoencoder-like mapping in situations where the input images are highly correlated with the desired outputs may be wasteful especially for large and deep networks.
It may be one of the main reasons why Dong \textit{et al.}~\cite{Dong2015} were not able to scale up their networks and why they required approximately $10^7$ iterations to train their AR-CNN.
Similar problems were reported by Kim \textit{et al.}~\cite{Kim2015}.

\begin{figure}[b]
	\centering
	\includegraphics[width=\linewidth]{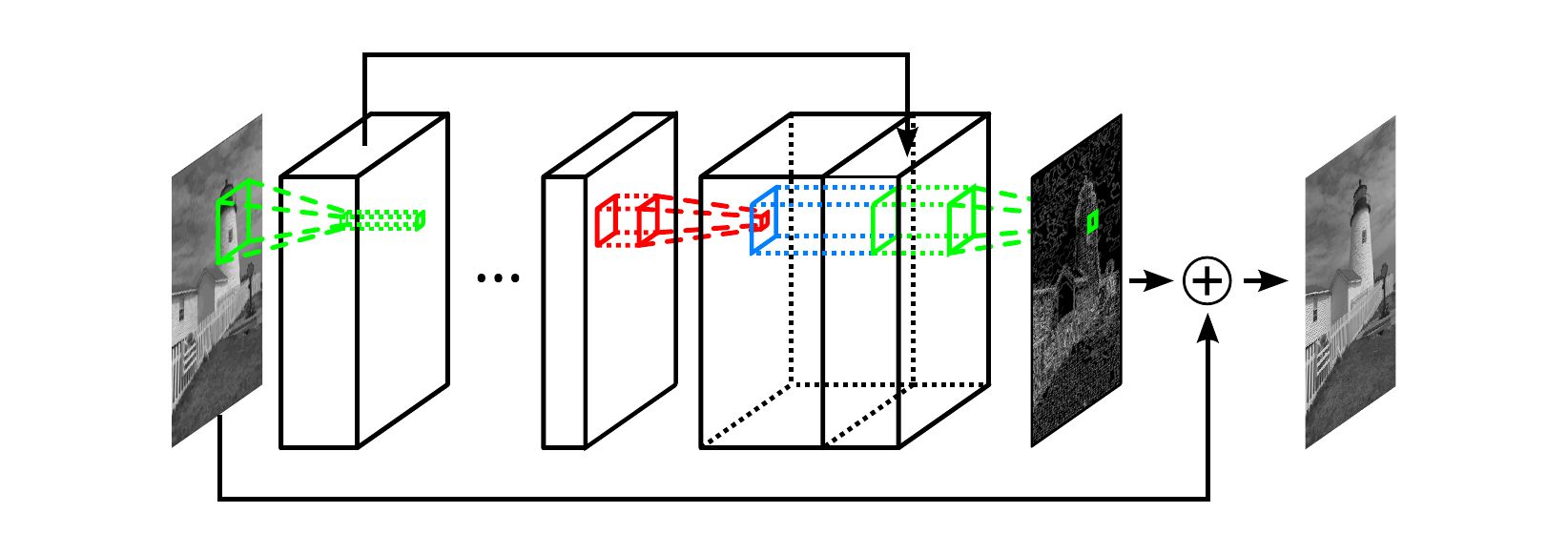}
	\caption{Illustration of a network with skip architecture and residual loss.}
	\label{fig:schema-skip-residual}
\end{figure}

\medskip
\subparagraph{Residual objective.}
The residual objective was originally introduced for super-resolution~\cite{Kim2015} where the input and output images are highly correlated.
Instead of learning to predict the output image, the network in residual learning learns the changes which should be applied to the input image -- it predicts the residual image ${r = y - \hat{x}}$ between the distorted $y$ and latent high-quality image $\hat{x}$.
The residual learning scheme is depicted in Figure \ref{fig:schema-skip-residual}.
Kim \textit{et al.}~\cite{Kim2015} were able to speed up the training by large factor of (up to $10^4\times$) with residual learning and it allowed them to learn much deeper networks -- 20 layers vs. 3 in \cite{Dong2014} and 4 in \cite{Dong2015}.

\begin{figure}[b]
	\centering
	\begin{scriptsize}
		\def\svgwidth{\linewidth}
		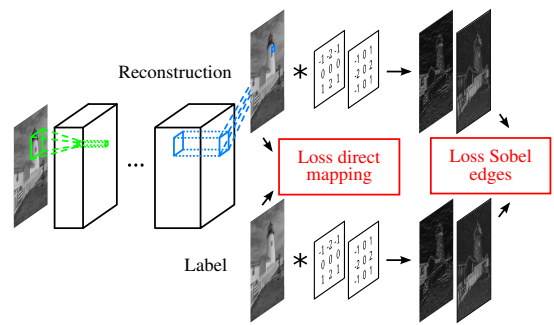
	\end{scriptsize}
	\caption{Illustration of a network with edge preserving loss.}
	\label{fig:schema-sobel}
\end{figure}

\medskip
\subparagraph{Edge emphasized objective.}
Mean square error used in many image restoration methods does not necessarily correlate well with the image quality perceived by humans.
With convolutional networks, it is relatively easy to use more perceptually valid error measures as objective functions, as long as they can be efficiently differentiated (e.g. SSIM).
We decided to add partial first derivatives of the image to the loss function in a form of vertical and horizontal Sobel filters.
This is achieved by adding the objective function computed on image derivative calculated by Sobel filter $G$ as
\begin{align}
	\frac{1}{n} \sum_{i=1}^{n} \left\lVert G \ast F(y_{i}) - G \ast x_{i} \right\rVert_{2}^{2}.
	\label{eq:objective-edge-preserving}
\end{align}
Our assumption is that the addition of the first derivatives should force the network to focus specifically on high frequency structures such as edges, ringing artifacts, and block artifacts and it could lead to perceptually better reconstructions.
The combined edge emphasized loss can be easily implemented in all existing convolutional network frameworks by defining the derivative Sobel filters as a convolutional layer with predefined fixed filters. The network utilizing such objective function is shown in Figure \ref{fig:schema-sobel}.

\medskip
\subparagraph{Symmetric weight initialization.}
Weights in convolutional networks are usually initialized by sampling from some simple distribution (e.g. Gaussian or uniform) with mean equal to 0.
The zero mean is desirable as it prevents mean offsets of activations to propagate through the layers.
In case the mean was not zero, any mean offset in input values would result in non-zero mean of output activations which could force the ReLU non-linearities to get fully stuck either in the positive linear interval or, even worse, in the negative interval where gradients are not propagated rendering the unit useless.

\begin{table*}
	\centering
	\begin{tabu} to \linewidth {l|X[c]X[c]X[c]X[c]X[c]X[c]X[c]X[c]}
		\toprule
		Layer 			& $1$ & $2$ & $3$ & $4(+1)$ & $5$ & $6(+1)$ & $7$ & $8$ \\
		\midrule
		Filter size		& $11\times11$ & $3\times3$ & $3\times3$ & $3\times3$ & $1\times1$ & $5\times5$ & $1\times1$ & $5\times5$ \\
		Channels	    & $32$ & $64$ & $64$ & $64(+32)$ & $64$ & $64(+32)$ & $128$ & $1$ \\
		\bottomrule
	\end{tabu}
	\caption{
		L8 architecture -- filter size and number of channels for each layer.
	}
	\label{tab:L8_architecture}
\end{table*}

Although the weights are sampled from a distribution with zero mean, the means of individual convolutional filters are not zero due to the fact that they are a finite sample from the distribution.
These random offsets together with the positive offset of ReLU activations cause units in deeper layers to become more likely to be either permanently turned off or turned on, which increases sparsity of the activations and increases effective mean offsets of the deeper layers.
The result is that that majority of units in deep layers become almost useless right after the initialization.

Some activation normalization methods, such as "batch normalization"~\cite{Ioffe2015}, can eliminate the saturation problem, but the normalization introduces noise during training which is not desirable for image restoration networks.

We eliminate this problem by explicitly forcing individual filters to have zero mean during initialization.
Such initialization allows us to use significantly higher initial learning rates, especially together with residual learning, and it results in trained networks with significantly fewer saturated neurons.

We could explicitly force all filters to have zero mean during the whole training.
Such constraint almost entirely eliminates any potential for unit saturation, but it prevents networks to utilize the DC component of input signals.
Although we were able to achieve reasonably good results with this constraint in our preliminary experiments, we did not find it necessary and it was not used in the experiments presented in this paper.

\medskip
\subparagraph{Skip architecture.}
Deeper networks may have problems with exploding and vanishing gradients and they may take a long time to learn to efficiently propagate information through large number of layers.
The problems with the gradients can be eliminated by proper initialization~\cite{Glorot2010}.
The problems with propagating information through many layers can be alleviated by bypassing some layers~\cite{Long2015} or by letting layers to learn residual of their inputs~\cite{He2015}.
The skip architecture with the residual objective function is shown in Figure \ref{fig:schema-skip-residual}.

We employ a skip architecture similarly to Long \textit{et al.}~\cite{Long2015}.
We feed activations of the first convolutional layer to some deeper layers bypassing the layers in-between.
Unlike Long \textit{et al.}~\cite{Long2015} who add the activations together, we concatenate them.
The goal of the skip architecture is to allow the network to pass geometric information easily from the input to the output, and to allow for more complex reasoning about the image content in the middle layers (e.g. what is an artifact and what local context information should be used to repair the artifacts.

\begin{table}
	\centering
	\begin{tabu} to \linewidth {l|X[c]X[c]X[c]X[c]}
		\toprule
		Layer 			&          $1$ &        $2$ &        $3$ & $4$ \\
		\midrule
		Filter size		& $11\times11$ & $3\times3$ & $3\times3$ & $5\times5$ \\
		Channels	    &         $48$ &       $64$ &       $64$ & $1$ \\
		\bottomrule
	\end{tabu}
	\caption{
		L4 architecture -- filter size and number of channels for each layer.
	}
	\label{tab:L4_architecture}
\end{table}

\medskip
\subparagraph{Network architectures.}
This paper presents two different FCN architectures which use only convolutional units and ReLU non-linearities.
The first architecture denoted as L4 is relatively small with four layers defined in Table~\ref{tab:L4_architecture}.
The second network, denoted as L8, has eight layers and it utilizes the skip architecture by concatenating activation of the first layer with activations of the fourth and sixth layers.
The exact definition of L8 is in Table~\ref{tab:L8_architecture}.
The receptive fields of L4 and L8 are $19\times19$ and $25\times25$, respectively.

\section{Experimental Results}
\label{sec:Experimental-results}

All the experiments were computed on images from BSDS500~\cite{BSDS500} and LIVE1~\cite{LIVE1} datasets.
The networks were trained solely on the merged train and test part of BSDS500 which contain 400 images.
The images were transformed to gray-scale using the YCbCr color model by keeping the luma component -- Y only.
Although the networks can process color images, we evaluate on gray-scale images because we focus on the ringing and blocking artifacts and not on the chromatic distortions. 
The gray-scale images were compressed with the MATLAB JPEG encoder into six disjoint sets according the JPEG quality.
Specifically, we use images compressed with the quality 10, 20, 30, 40, 50, and 60.

The networks were evaluated on the validation set from BSDS500 which includes 100 high quality compressed images and on the LIVE1 dataset containing 29 color images (uncompressed BMP format).
All the evaluation images were transformed to gray-scale the same way as the training images and compressed using the same encoder.

\begin{table*}
	\centering
    \begin{tabu} to \linewidth {l|X[r]X[r]X[r]|X[r]X[r]X[r]}
    		\toprule
            & \multicolumn{3}{c|}{Q10} & \multicolumn{3}{c}{Q20} \\
            \tabucline{2-}\vphantom{\v{M}}
            method      & PSNR & PSNR-B & SSIM & PSNR & PSNR-B & SSIM \\
            \midrule
            distorted   & 27.77 &    25.33 &  0.791 &    30.07 &    27.57 &  0.868 \\
            spp         & 28.37 &    27.77 &  0.806 &    30.49 &    29.22 &  0.877 \\
            SA-DCT      & 28.65 &    28.01 &  0.809 &    30.81 &    29.82 &  0.878 \\
            AR-CNN      & 28.98 &    28.70 &  0.822 &    31.29 &    30.76 &  0.887 \\
            L4 Residual & \textbf{29.08} & \textbf{28.71} & \textbf{0.824} & 31.42 &    30.83 &  0.890 \\
            L8 Residual & -- & -- & -- & \textbf{31.51} & \textbf{30.92} & \textbf{0.891} \\
            \bottomrule
    \end{tabu}
    \medskip
    \caption{
    	Image reconstruction quality on LIVE1 validation dataset for JPEG quality 10 and 20.
    }
    \label{tab:results-dataset-LIVE1}
\end{table*}

\begin{table*}
	\centering
    \begin{tabu} to \linewidth {l|X[r]X[r]X[r]|X[r]X[r]X[r]}
    		\toprule
            & \multicolumn{3}{c|}{Q10} & \multicolumn{3}{c}{Q20} \\
            \tabucline{2-}\vphantom{\v{M}}
            method      & PSNR & PSNR-B & SSIM & PSNR & PSNR-B & SSIM \\
            \midrule
            distorted   & 27.58 &    24.97 &  0.769      &    29.72 &    26.97 &  0.852 \\
            spp         & 28.13 &    27.49 &  0.782      &    30.11 &    28.68 &  0.859 \\
            AR-CNN      & 28.74 & \textbf{28.38} & 0.796 &    30.80 &    30.08 &  0.868 \\
            L4 Residual & \textbf{28.75} & 28.29 & \textbf{0.800} &    30.90 &    30.13 &  0.871 \\
            L8 Residual & -- & -- & -- &    \textbf{30.99} & \textbf{30.19} & \textbf{0.872} \\
            \bottomrule
    \end{tabu}
    \medskip
    \caption{
    	Image reconstruction quality on BSDS500 validation dataset for JPEG quality 10 and 20.
    }
    \label{tab:results-dataset-BSR}
\end{table*}

\begin{figure*}
	\centering
    \begin{subfigure}[b]{.24\linewidth}
		\includegraphics[width=\linewidth]{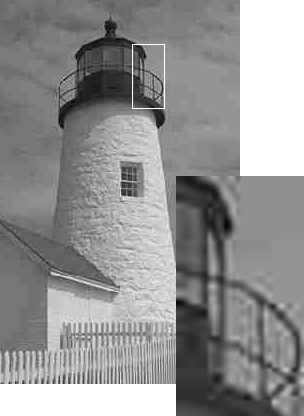}
        \caption{distorted}
        \bigskip
        \label{fig:wide-visual-lighthouse3-distorted}
    \end{subfigure}%
    \hspace*{\fill}
    \begin{subfigure}[b]{.24\linewidth}
		\includegraphics[width=\linewidth]{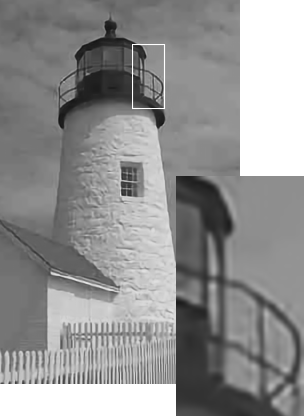}
        \caption{AR-CNN}
        \bigskip
        \label{fig:wide-visual-lighthouse3-arcnn}
    \end{subfigure}%
    \hspace*{\fill}
    \begin{subfigure}[b]{.24\linewidth}
		\includegraphics[width=\linewidth]{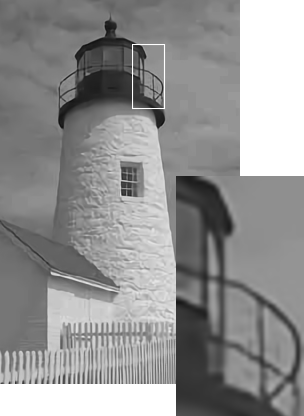}
        \caption{L08}
        \bigskip
        \label{fig:wide-visual-lighthouse3-L08}
    \end{subfigure}%
    \hspace*{\fill}
    \begin{subfigure}[b]{.24\linewidth}
		\includegraphics[width=\linewidth]{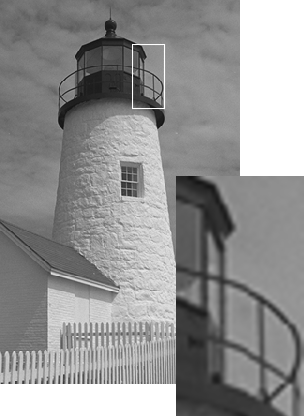}
        \caption{original}
        \bigskip
        \label{fig:wide-visual-lighthouse3-original}
    \end{subfigure}%
    \caption{Illustrative comparison of reconstruction quality on lighthouse3 image from LIVE1 dataset, for JPEG quality~20.}
    \label{fig:wide-visual-lighthouse3}
\end{figure*}

\begin{figure*}
	\centering
    \begin{subfigure}[b]{.33\linewidth}
		\includegraphics[width=\linewidth]{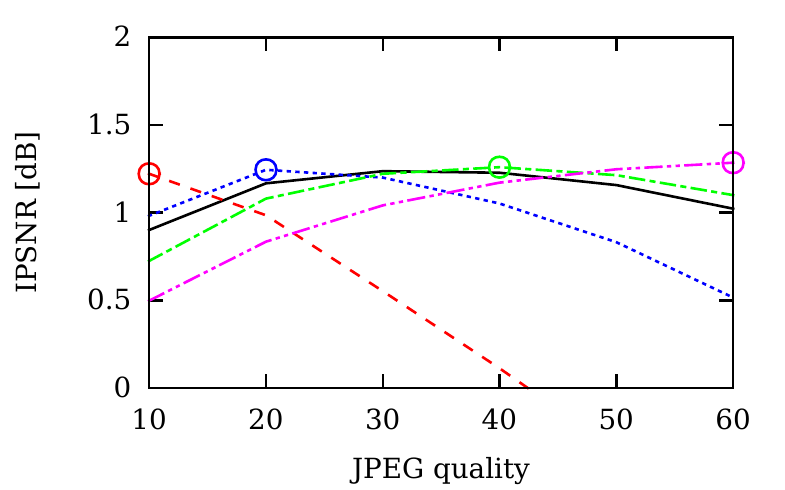}
        \caption{Normal}
        \label{fig:wide-results-Normal}
    \end{subfigure}%
    \begin{subfigure}[b]{.33\linewidth}
		\includegraphics[width=\linewidth]{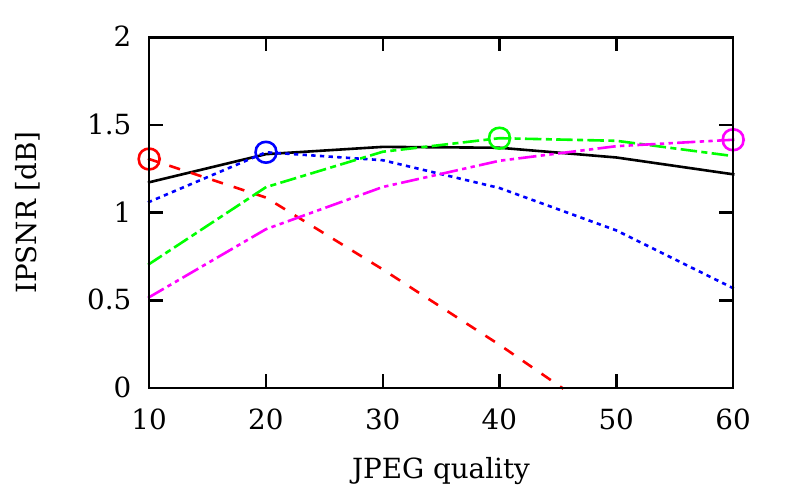}
        \caption{Residual}
        \label{fig:wide-results-Residual}
    \end{subfigure}%
    \begin{subfigure}[b]{.33\linewidth}
		\includegraphics[width=\linewidth]{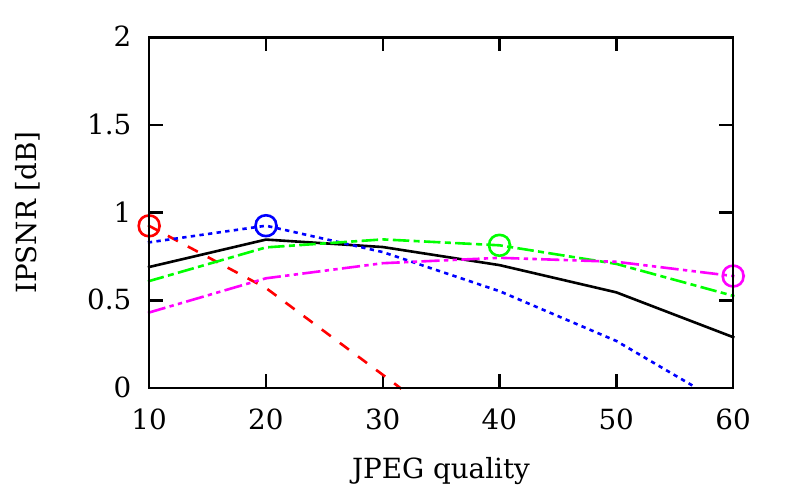}
        \caption{Sobel}
        \label{fig:wide-results-Sobel}
    \end{subfigure}\\
    \bigskip
    \includegraphics[width=.42\linewidth]{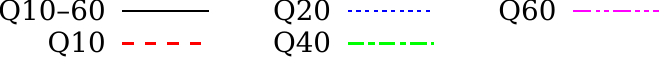}\\
    \bigskip
    \caption{Generalization ability of L4 networks trained with Normal, Residual, and Edge preserving objectives for different JPEG quality levels.}
    \label{fig:wide-results}
\end{figure*}

Several metrics for objectively assessing perceptual quality of images exist.
We use PSNR, PSNR-B, and SSIM.
Generally, the most commonly used quality metric is the mean squared error (MSE).
This quantity is computed by averaging squared intensity differences of the distorted image and the reference image.
The quantity is often expressed in a logarithmic scale as the peak signal-to-noise ratio (PSNR).
Unfortunately, PSNR and MSE are not necessarily correlated well with perceptual quality.
The structural similarity index (SSIM)~\cite{Wang2004} that compares local patterns of pixel intensities should be better correlated with perceptual quality.
Since we focus on JPEG artifacts which include blocking artifacts, a block-sensitive metric referred to as the \mbox{PSNR-B}~\cite{Yim2011} should provide additional insights.
\mbox{PSNR-B} modifies the original PSNR by including an additional blocking effect factor (BEF).
Some experiments report IPSNR which is a PSNR increase compared to PSNR of the degraded image.
IPSNR is more stable across different dataset and it directly reflects the quality improvement.

We compare our results to AR-CNN~\cite{Dong2015}, to the widely regarded deblocking oriented SA-DCT~\cite{Foi2006, Foi2007}, and to a simple postprocessing filter spp included in the FFmpeg framework~\cite{Nosratinia1999}.
While L4 was used in most experiments and it was trained for various compression quality levels, L8 was trained only for quality 20.
If not stated otherwise, the residual version of networks was used.

The L4 and L8 networks were trained on mini-batches of 64 $64\times64$ patches and 4 $128\times128$ patches respectively.
The patches were randomly sampled from training images.
The number of training iterations was fixed to 250\;K which is significantly less compared to AR-CNN's $10^7$ iterations.
The learning rate was scaled down by factor of 2 every 50\;K iterations.
The networks were initialized by the Xavier initialization~\cite{Glorot2010} in the first three layers, and a Gaussian initialization with lower variation was used in the final layer.
The learning rate of the last layer was set ten times smaller than for the other layers.

\medskip
\subparagraph{Artifacts reduction quality.}
The results of artifacts removal on LIVE1 dataset with JPEG quality 10 and 20 are shown in Table~\ref{tab:results-dataset-LIVE1}.
The results on the BSDS500 validation set are presented in Table~\ref{tab:results-dataset-BSR}.
L8 outperforms all other methods with significantly higher scores in all three quality measures.
L4 which performs worse compared to L8, still surpasses the other methods in most cases even though it is much small and computationally efficient compared to L8.
Examples of resulting images are presented in Figure~\ref{fig:wide-visual-lighthouse3}.

\medskip
\subparagraph{JPEG quality generalization.}
We evaluated the ability of the trained networks to generalize to a different compression quality by training L4 on one quality and evaluating on other qualities (L4Q10 trained for quality 10, L4Q20 for quality 20, etc up to L4Q60).
To asses the ability of CNNs to handle multiple compression qualities in a single model, we trained a single L4 network on all the qualities together (L4Q10-Q60).
The results in Figure~\ref{fig:wide-results} show that L4Q10-Q60 provides stable results across the quality range.
However, the quality-specific networks perform better for their respective qualities.
The quality-specific networks generalize only to similar qualities.
In practice, a single network should easily be able to handle smaller quality ranges (e.g. 10--20 quality points wide) when trained on data from the whole range.

\begin{figure}[t]
	\centering
    \includegraphics[width=\linewidth]{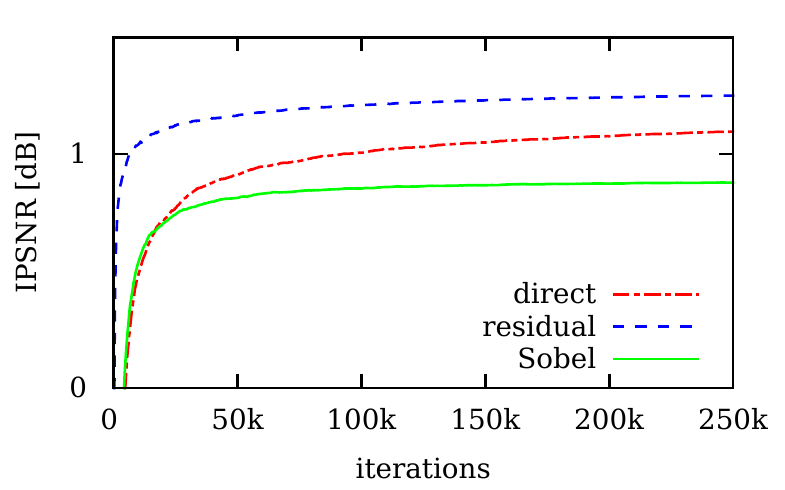}
	\caption{Training development of L4 with different training objectives.}
    \label{fig:train-ipsnr}
\end{figure}

\medskip
\subparagraph{Impact of learning objective.}
We compare L4 networks trained for direct mapping, residual, and edge preserving loss.
Although the architecture and initialization of all the L4 networks were the same, we had to select suitable learning rates (lr) and weight decay coefficients (wd) by performing grid search for each learning objective separately.
The chosen values are for direct mapping lr 0.4, wd $5 \times 10^{-7}$, for residual learning lr 8, wd $5 \times 10^{-7}$, and for edge preserving objective lr 0.05, wd $5 \times 10^{-4}$.%
\footnote{In our experiments, the loss was normalized by the number of output pixels.
This scaling influences the scale of gradients and results in relatively high learning rates and low weight decay coefficients.}
The values were chosen on JPEG quality 10 and they were used for all other qualities.

\begin{figure}
	\centering
    Normal\\
    \smallskip
    \begin{subfigure}[b]{.3\linewidth}
    	\includegraphics[width=\linewidth]{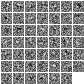}
        \caption{20k}
        \label{fig:filters-normal-020000}
    \end{subfigure}%
    \hspace*{\fill}
    \begin{subfigure}[b]{.3\linewidth}
    	\includegraphics[width=\linewidth]{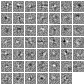}
        \caption{100k}
        \label{fig:filters-normal-100000}
    \end{subfigure}%
    \hspace*{\fill}
    \begin{subfigure}[b]{.3\linewidth}
    	\includegraphics[width=\linewidth]{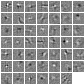}
        \caption{250k}
        \label{fig:filters-normal-247500}
    \end{subfigure}\\
    \medskip
    Residual\\
    \smallskip
    \begin{subfigure}[b]{.3\linewidth}
    	\includegraphics[width=\linewidth]{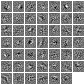}
        \caption{5k}
        \label{fig:filters-residual-005000}
    \end{subfigure}%
    \hspace*{\fill}
    \begin{subfigure}[b]{.3\linewidth}
    	\includegraphics[width=\linewidth]{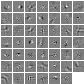}
        \caption{20k}
        \label{fig:filters-residual-020000}
    \end{subfigure}%
    \hspace*{\fill}
    \begin{subfigure}[b]{.3\linewidth}
    	\includegraphics[width=\linewidth]{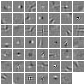}
        \caption{250k}
        \label{fig:filters-residual-247500}
    \end{subfigure}\\
    \medskip
    Sobel\\
    \smallskip
    \begin{subfigure}[b]{.3\linewidth}
    	\includegraphics[width=\linewidth]{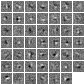}
        \caption{2.5k}
        \label{fig:filters-sobel-002500}
    \end{subfigure}%
    \hspace*{\fill}
    \begin{subfigure}[b]{.3\linewidth}
    	\includegraphics[width=\linewidth]{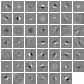}
        \caption{20k}
        \label{fig:filters-sobel-020000}
    \end{subfigure}%
    \hspace*{\fill}
    \begin{subfigure}[b]{.3\linewidth}
    	\includegraphics[width=\linewidth]{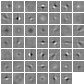}
        \caption{250k}
        \label{fig:filters-sobel-247500}
    \end{subfigure}\\
    \smallskip
	\caption{
    	Filters from the first layer of L4 networks with normal/residual/Sobel (edge preserving) objective at different stages of training.
        Iterations are showed below the images.
    }
    \label{fig:filters}
\end{figure}

The progress of learning is shown in Figure~\ref{fig:train-ipsnr}. 
The residual network converges much faster compared to the direct mapping network. 
The results on LIVE1 measured by PSNR, \mbox{PSNR-B} and SSIM are in Table~\ref{tab:learn-target-results}.

\begin{table}[b]
	\centering
	\begin{tabu} to \linewidth {l|X[c]X[c]X[c]}
		\toprule
		Objective        &            PSNR &  \mbox{PSNR-B} &   SSIM \\
		\midrule
		Distorted              &           27.58 &          24.97 &  0.769 \\
		Direct mapping         &           28.99 &          28.66 &  0.820 \\
		Edge preserving        &           28.69 &          28.40 &  0.813 \\
		Residual learn.        &  \textbf{29.08} & \textbf{28.71} &  \textbf{0.824} \\
		\bottomrule
	\end{tabu}
	\caption{
		Results of L4 networks with different objectives on LIVE1 dataset with quality 10.
	}
	\label{tab:learn-target-results}
\end{table}

Figure~\ref{fig:filters} shows 1st layer filters of the networks during different stages of training. 
All the networks formed reasonable-looking filters.
The residual network formed more complex higher frequency filters compared to the other networks. 
The edge preserving network learned a number of low-pass filters which are probably needed to transfer the general image appearance through the network -- these filters are missing in the residual network.
The filters of the normal direct mapping network remain noisy, which could be due to different weight decay coefficient the low learning rate, or their combination.

The results show that the residual learning is beneficial for JPEG artifact reduction in terms of resulting reconstruction quality and training speed.
On the other hand, the edge preserving objective does not improve resulting quality noticeably in the case of L4.

\begin{figure}
	\centering
    \begin{subfigure}[b]{\linewidth}
    	\includegraphics[width=\linewidth]{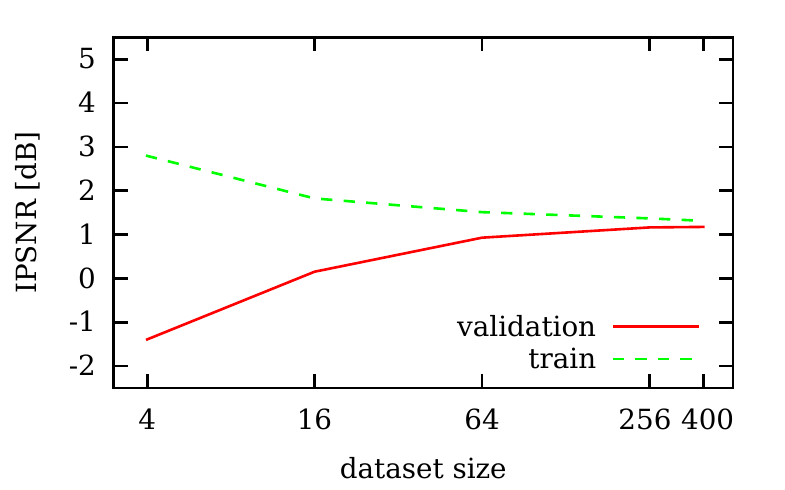}
        \caption{L4}
        \label{fig:dataset-size-L4}
    \end{subfigure}\\
    \begin{subfigure}[b]{\linewidth}
    	\includegraphics[width=\linewidth]{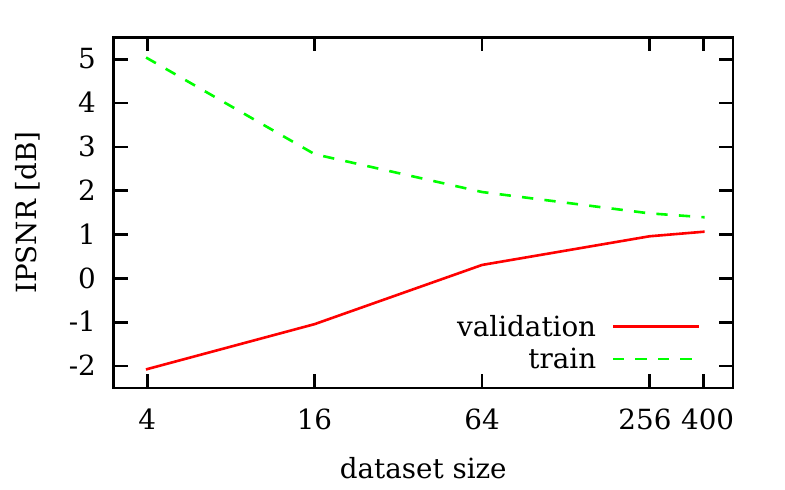}
        \caption{L8}
        \label{fig:dataset-size-L8}
    \end{subfigure}
    \smallskip
	\caption{Generalization for different sized train set.}
    \label{fig:dataset-size}
\end{figure}

\subparagraph{Dataset size.}
The quality of reconstruction achieved by larger networks may suffer due to inadequate size of a training set.
In order to asses how the L4 and L8 behave with respect to training set size, we trained the residual versions of the networks on 4, 16, 64, 256, and 400 images from the training set.
The L4 and L8 networks contain approx 70\;K and 220\;K learnable parameters respectively which suggests that L8 should require larger training set for the same generalization.
Figure~\ref{fig:dataset-size} shows results on the different training sets and corresponding results on the independent test set.
Both networks clearly overfit on the smaller datasets.
L8 overfits significantly more and it would require more images to reach proper generalization, while L4 seems to reach perfect generalization already on the relatively small dataset of 400 images.

\subparagraph{Speed.}
Using cuDNN v3 implementation of convolutions on GeForce GTX 780, we were able to process 1\;Mpx images in 220\;ms with network L4 and in 1052\;ms with L8.
The L4 and L8 networks require approximately 140\;K and 440\;K floating point operations per pixel, respectively.%
\footnote{The networks, processed images, and implementations are available at\\\scriptsize\url{http://www.fit.vutbr.cz/~ihradis/CNN-Deblur/}.}

\section{Conclusions}
\label{sec:Conclusions}

In this work, we show that it is possible to train large and deep networks for JPEG artifacts removal which outperform previous state-of-the-art results of smaller networks.
We combine the residual learning by Kim \textit{et al.}~\cite{Kim2015}, skip architecture~\cite{Long2015}, and symmetric weight initialization which allowed us to successfully train networks with 8 layers.

We compare networks with three different objectives -- direct mapping, residual learning, and edge preserving.
The best reconstruction results are provided by the residual learning.

We further investigate the network ability to generalize across different compression JPEG quality levels.
Our results show that it is possible to use one network trained for several qualities as an acceptable trade-off.

Finally, we evaluate generalization of the networks with respect to training set size. 
The results suggest that small networks similar to L4 (20\;K parameters) can be safely trained on the BSD dataset.
However, the generalization of L8 (100\;K parameters) and larger networks is not guaranteed on this small dataset and a larger common dataset should be compiled to allow fair and consistent evaluation in the future.


In a future work, we intend to apply convolutional networks to other compression methods, for example, JPEG 2000, JPEG XR, or WebP.
Next, we would like to train convolutional networks to reconstruct images directly from the JPEG coefficients which should provide the networks with significant clues as to which image elements are and which are not artifacts.
The receptive field even of the L8 network is still relatively small and we expect that it should be possible to reach higher reconstruction quality by increasing the receptive field or by providing context information by other means.

\paragraph{Acknowledgements}
This work has been supported by the ARTEMIS joint undertaking under grant agreement ALMARVI (no. 621439), the Technology Agency of the Czech Republic (TA CR) Competence Centres project V3C -- Visual Computing Competence Center (no. TE01020415) and the Ministry of Education, Youth and Sports from the National Programme of Sustainability (NPU II) project IT4Innovations excellence in science (no. LQ1602).

\medskip\medskip
\bibliographystyle{abbrvnat}
\bibliography{sources}

\end{document}